\documentclass{article}

\usepackage{PRIMEarxiv}

\usepackage[utf8]{inputenc} 
\usepackage[T1]{fontenc}    
\usepackage{hyperref}       
\usepackage{url}            
\usepackage{booktabs}       
\usepackage{amsfonts}       
\usepackage{nicefrac}       
\usepackage{microtype}      
\usepackage{lipsum}
\usepackage{fancyhdr}       
\usepackage{graphicx}       
\graphicspath{{media/}}     
\usepackage{amsmath} 
\usepackage[ruled,linesnumbered]{algorithm2e}
\usepackage{placeins}
\title{CCRA: Optimizing Vision-Language Consistency via Cross-Layer Regional Attention Alignment
}

\author{
  Yifan Wang\thanks{These authors contributed equally.} \\
  School of Medicine, \\
  The Chinese University of Hong Kong, Shenzhen \\
  \texttt{(yifanwang13@link.cuhk.edu.cn)} \\
  \and
  Hongfeng Ai\footnotemark[1] \\
  School of Medicine, \\
  The Chinese University of Hong Kong, Shenzhen \\
  \texttt{(aihongfeng@cuhk.edu.cn)} \\
  \and
  Quangao Liu\footnotemark[1] \\
  Shenyang Institute of Automation, \\
  Chinese Academy of Sciences \\
  \texttt{liuquangao@sia.cn} \\
  \and
  Maowei Jiang\footnotemark[1] \\
  Shenzhen International Graduate School, \\
  Tsinghua University \\
  \texttt{(jiangmaowei@sia.cn)} \\
  \and
  Ruiyuan Kang\footnotemark[1] \\
  Wave and Machine Intelligence Department, \\
  Technology Innovation Institute \\
  \texttt{ruiyuan.kang@tii.ae} \\
  \and
  Ruiqi Li \\
  University of the Chinese Academy of Sciences \\
  \texttt{(liruiqi1@sia.cn)} \\
  \and
  Jiahua Dong \\
  Mohamed bin Zayed University of \\
  Artificial Intelligence \\
  \texttt{dongjiahua1995@gmail.com} \\
  \and
  Mengting Xiao \\
  McGill University \\
  \texttt{mengting.xiao@mail.mcgill.ca} \\
  \and
  Cheng Jiang\thanks{Corresponding author.} \\
  School of Medicine, \\
  The Chinese University of Hong Kong, Shenzhen \\
  \texttt{jiangcheng@cuhk.edu.cn} \\
  \and
  Chenzhong Li\footnotemark[2] \\
  School of Medicine, \\
  The Chinese University of Hong Kong, Shenzhen \\
  \texttt{lichenzhong@cuhk.edu.cn} \\
}

\begin{document}
\maketitle

\begin{abstract}
Vision Language Models (VLMs) face challenges in effectively coordinating diverse attention mechanisms for cross-modal embedding learning, leading to mismatched attention and suboptimal performance. We propose Consistent Cross-layer Regional Alignment (CCRA), which introduces Layer-Patch-wise Cross Attention (LPWCA) to capture fine-grained regional-semantic correlations by jointly weighting patch and layer-wise embedding, and Progressive Attention Integration (PAI) that systematically coordinates LPWCA, layer-wise, and patch-wise attention mechanisms in sequence. This progressive design ensures consistency from semantic to regional levels while preventing attention drift and maximizing individual attention benefits. Experimental results on ten diverse vision-language benchmarks demonstrate that our CCRA-enhanced LLaVA-v1.5-7B model achieves state-of-the-art performance, outperforming all baseline methods with only 3.55M additional parameters, while providing enhanced interpretability through more regionally focused and semantically aligned attention patterns.
\end{abstract}

\keywords{VLM \and Cross-Attention \and Multimodal Alignment \and Layer-Patch-wise Attention \and Model Interpretability}

\section{Introduction}
Vision Language Models (VLMs) have fundamentally transformed visual question answering \cite{jia2024vqa2visualquestionanswering}, object detection \cite{li2023evaluatingobjecthallucinationlarge}, segmentation \cite{Khan_2022}, OCR \cite{singh2019textvqa}, etc. A key insight is that diverse tasks, represented by different text queries, require significantly different information from the given image—not only regarding which regions to focus on, but also which embedding layers should receive more attention when transferring semantic information \cite{lin2025healthgptmedicallargevisionlanguage}. This presents a fundamental challenge: how to optimize vision information extraction to better align with the specific needs of text queries for optimal performance.

Existing approaches to vision-language alignment fall into several categories. Some methods extract image embeddings from specific layers of the vision encoder and then perform Patch-Wise Cross Attention (PWCA) between textual and visual embeddings \cite{lee2021patch, luo2022frustratingly, alayrac2022flamingo,jiang2024timixtextawareimagemixing,jiang2024busefficienteffectivevisionlanguagepretraining, diao2025evev2}. However, diverse tasks often require a different emphasis on visual features at multiple semantic levels \cite{lin2025healthgptmedicallargevisionlanguage}. To address this limitation in vision-language alignment, other approaches employ Layer-Wise Cross Attention (LWCA) to assign importance weights across different layers \cite{sung2023ecoflap, li2025instructionguidedfusionmultilayervisual, kaduri2025s}. Most recently, Liu et al. \cite{liu2025laco} unified both PWCA and LWCA mechanisms under a single framework via MLP-based patch information compression, providing a more comprehensive consideration.

Despite these advances, a critical limitation persists: harmonic coordination between diverse attention mechanisms lacks effective organization, potentially leading to mismatched attention from different perspectives and resulting in suboptimal performance and poor interpretability.To address this limitation, we propose \textbf{Consistent Cross-layer Regional Alignment (CCRA)} with two key contributions:
\begin{enumerate}
    \item \textbf{Layer-Patch-Wise Cross Attention (LPWCA)}: Beyond existing LWCA and PWCA, we introduce LPWCA to capture fine-grained regional-semantic correlations, enabling superior performance across diverse tasks.
    \item \textbf{Progressive Attention Integration (PAI)}: We systematically integrate all three attention mechanisms through progressively operating LPWCA, optimized Gaussian-smoothed LWPA and finally PWCA.This design maximizes the benefits of individual attention mechanisms while ensuring consistency in both semantic and regional levels, enhancing both performance and interpretability.
\end{enumerate}

To demonstrate CCRA's effectiveness in improving generalization performance and interpretability, we evaluate our CCRA-enhanced LLaVA-v1.5-7B model on diverse vision tasks and visualize attention patterns through feature heatmaps. Our results demonstrate that the proposed model outperforms all baseline methods across contrastively different tasks with diverse task queries. Meanwhile, the feature heatmaps visualize the adaptivity and consistency of feature attention, which supports the superior performance of VLM across diverse tasks, and also provide more interpretable visual representations of feature importance compared to existing approaches.

\section{Related work}
\subsection{VLM without Vision-Language Alignment}

Conventional VLM often decoupled visual and textual embeddings process, e.g., LLaVA,  MiniGPT-v2, and LLaMA-Adapter-v2~\cite{chen2023minigpt,gao2023llama, liu2024improved}. They often extract single-layer embedding from visual encoder, and feed with textual embedding to pre-trained encoders such as CLIP \cite{jiang2023clip}. However, diverse tasks often require visual features from a broader range of semantic levels \cite{kaduri2024whatsimagedeepdivevision,iana2024peelinglayersindepthevaluation}.
Accordingly, recent advances leverage cross-layer visual features for comprehensive representations \cite{chen2024evlmefficientvisionlanguagemodel,cao2024mmfuser,chen2024lion,yao2024dense}. These approaches capture both low-level details from early layers and high-level semantics from deeper layers.
To reduce feature redundancy and noise, these methods also involved similarity-based \cite{raghu2021vision,yao2024dense,sun2025transformer} and proportion-based \cite{cao2024mmfuser,chen2024lion,chen2024evlmefficientvisionlanguagemodel} layer feature selection has been explored. However, these methods operate independently of textual input, failing to consider that different tasks have varying visual requirements \cite{lin2025healthgptmedicallargevisionlanguage}. Early layers handle color and many spatial tasks such as counting or localization well \cite{chen2025rethinking, yao2024dense}. OCR is also sensitive to visual details in the shallow layers, and insufficient low-level information may lead to recognition errors \cite{cao2024mmfuser}. By contrast, high‑level semantic reasoning, long‑horizon action understanding, and knowledge‑intensive question answering rely on the deepest visual representations \cite{li2025instruction}. Such methods do not consider text-image alignment, leading to suboptimal VLM performance.


\subsection{VLM with Vision-Language Alignment}
Recent advances in vision-language alignment have explored various mechanisms to bridge textual semantics with visual representations. One line of research emphasizes PWCA, where image embeddings extracted from specific layers of the vision encoder are aligned with textual queries through cross-attention \cite{lee2021patch, luo2022frustratingly, alayrac2022flamingo, jiang2024timixtextawareimagemixing, jiang2024busefficienteffectivevisionlanguagepretraining, diao2025evev2}. This approach enhances fine-grained regional control and enriches visual representation, making it particularly effective for tasks requiring precise regional alignment \cite{yue2024mmmu}. Another line focuses on LWCA, which leverages cross-layer feature fusion to adaptively weight different visual semantic levels based on textual context \cite{sung2023ecoflap, li2025instructionguidedfusionmultilayervisual, kaduri2025s}. This facilitates better semantic alignment and improves generalization across diverse tasks.

Despite their respective strengths, LWCA and PWCA are typically designed independently, often lacking consistency between regional and semantic focus. This decoupled design leads to attention drift, where attention across layers inconsistently shifts regions of focus—undermining stable alignment and interpretability \cite{li2025instruction}. Moreover, relying solely on one form of attention neglects the relative importance between regional location and semantic depth, limiting the model’s ability to effectively optimize vision-language features.

To address these limitations, recent works have attempted to combine patch-wise and LWCAs. For example, Liu et al. \cite{liu2025laco} proposed a unified framework that compresses patch-level information via MLPs and integrates it with cross-layer attention, offering a more holistic alignment. However, rigid coordination may lead to inorganic coordination between different attentions, and provide suboptimal performance on complex multimodal tasks \cite{nam2017dual, liu2025laco}.

\section{Methodology}
\begin{figure*}[h]
\centering
\includegraphics[width=0.8\textwidth]{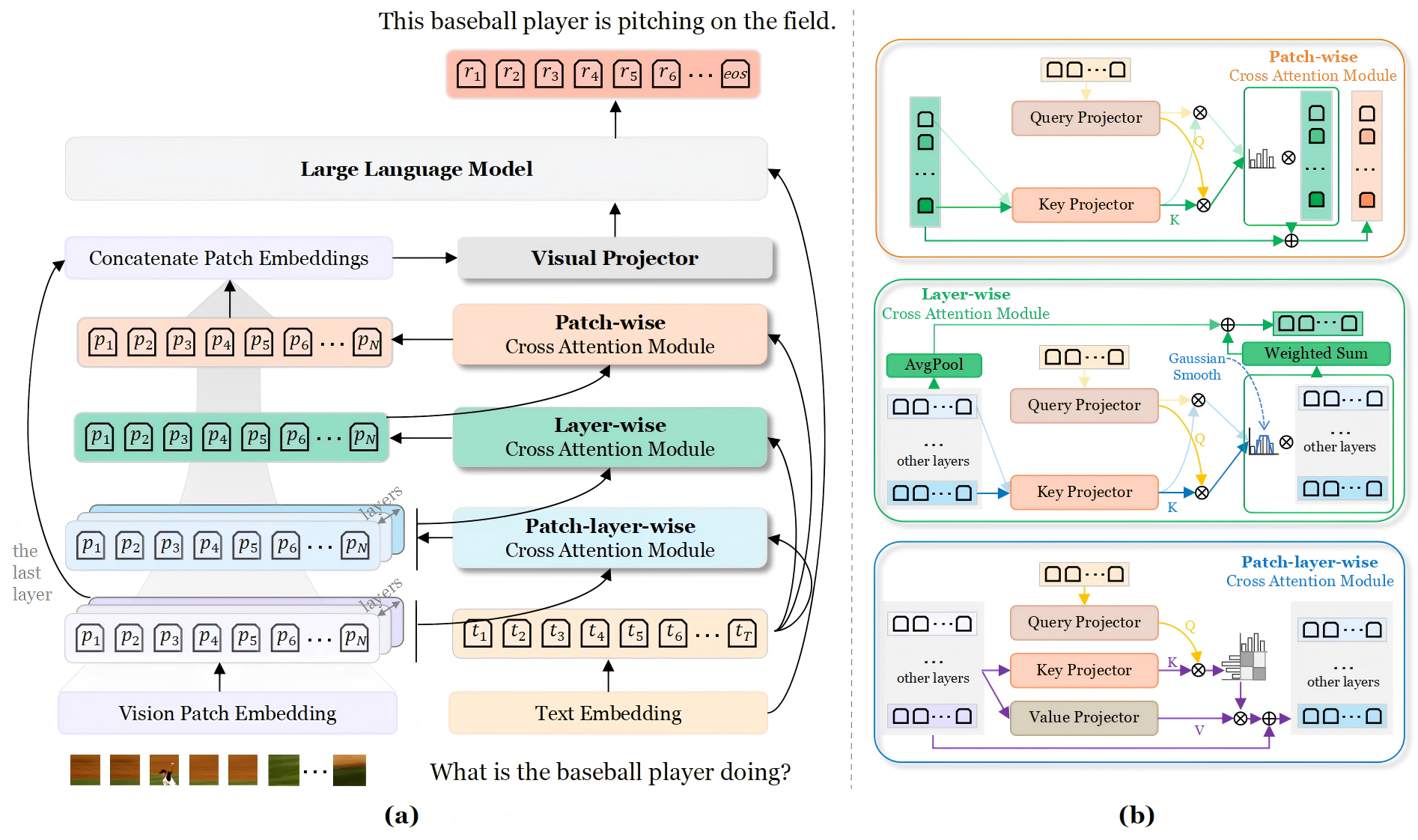}
\caption{(a) An overview of our VLM with Consistent Cross-layer Regional Alignment. Patch-Layer-Wise Cross Attention (PLWCA), Layer-Wise Cross-Attention(LWCA), and Patch-Wise Cross Attention(PWCA) are progressively used to align both textual embedding and visual embedding gradually for optimal task-oriented performance (b) The detailed illustration of PLWCA, LWCA and PWCA, where PLWCA provides joint and global optimization between regional and semantic information; the optimized Gaussian-smoothed LWCA provides continuous attention along semantic aspect; the PWCA provides the consistency constraint along regional apsect. }
\label{fig:model_overview}
\end{figure*}

As discussed above, diverse attention mechanisms have their specific usages, but a mechanism is needed to harmoniously integrate all these text-image cross attentions to globally optimize VLM's performance across tasks. In addition, considering the need for human being's understanding, we also need to consider the feature interpretability as a further constraint. To reflect these considerations, we propose  \textbf{Consistent Cross-layer Regional Alignment (CCRA)}, a novel framework to unify diverse text-image cross attention under one umbrella for optimal task-oriented perforamnce, and also support consistent feature attention for interpretable understanding. CCRA differs from the previous work in the following two aspects, which is highlighted in Figure \ref{fig:model_overview}.
\begin{enumerate}
    \item \textbf{L}ayer-\textbf{P}atch-\textbf{W}ise \textbf{C}ross \textbf{A}ttention (LPWCA): we first introduced LPWCA, to complement exising LWCA andPWCA. By such a design, we connect the correlation between layer- and patch-wise information, thus provide a finer-grained feature control than merely LWCA and PWCA. (Sec. \ref{sec:layerpatch})
    \item \textbf{P}rogressive \textbf{A}ttention \textbf{I}ntegration (PAI): In addition to LPWCA, we also considered PWCA and a dedicatedly optimized LWCA for providing additional attention perspectives smoothly, and also constraining the global semantic consistency and regional consistency. We proposed PAI to harmoniously unify all three attention mechanism in sequence for optimal task-orientedperformance, and provide consistent semantic and regional feature attention for human beings' understanding. (Sec.\ref{sec:unify})
\end{enumerate}

\subsection{Layer-Patch-Wise Cross Attention}
\label{sec:layerpatch}
In addition to LWCA and PWCA, which provide regional and semantic importance, respectively, we complement them with the layer-patch-wise cross attention (LPWCA) as a fundamental operation, to reflect the global and joint importance on both aspects.  

To do so, the multi-layer visual features extracted from a visual encoder (e,g., CLIP ViT \cite{radford2021learning}):  ${\mathbf{F}^l_v \in \mathbb{R}^{N \times d}}$, where $l \in \{1, 2, \ldots, L\}$ is the layer index, and $N$, $d$ are the number of image patches and the feature dimension, respectively, are flattened into a unified patch-layer feature sequence $\mathbf{F}_{\text{stack}}$:

\begin{equation}
\mathbf{F}_{\text{stack}} = \left[\mathbf{F}^1_v; \mathbf{F}^2_v; \cdots; \mathbf{F}^L_v\right] \in \mathbb{R}^{L \times (N \times d)}.
\end{equation}
Through such a way, the hierarchical structure of feature space, which can be viewed from patch and layer perspective, are unified into the same space. 

Then in order to align with the textual query, we first process the query into a set of textual embeddings $\mathbf{F}_{t} \in \mathbb{R}^{T \times d}$. A self-attention module is first applied over $\mathbf{F}_{t}$ to compute token-contextualised importance scores $\boldsymbol{\alpha}_t \in \mathbb{R}^{T}$, which indicate the relative contribution of each token in guiding the visual alignment.
\begin{equation}
\boldsymbol{\alpha}_t = \text{Softmax}(\text{SelfAttention}(\mathbf{F}_{t})).
\end{equation}

Next, the textual embeddings $\mathbf{F}_{t}$ are projected into a query space $\mathbf{Q}(\mathbf{F}_t)$, while the stacked visual features $\mathbf{F}_{\text{stack}}$ are projected into a key space $\mathbf{K}(\mathbf{F}_{\text{stack}})$. The layer-patch attention scores $\mathbf{A}_{lp}\in \mathbb{R}^{T \times (L \times N)}$ can thus be computed:

\begin{equation}
\mathbf{A}_{lp} = \frac{1}{\sqrt{d}} \mathbf{Q}(\mathbf{F}_t) \mathbf{K}(\mathbf{F}_{\text{stack}})^{\top}.
\end{equation}

Then $\mathbf{A}_{lp}$ is aggregated across all textual tokens using the learned importance weights $\boldsymbol{\alpha}_t$ to form a unified attention map $\mathbf{W}_{lp} \in \mathbb{R}^{L \times N}$ over spatial patches and layers:

\begin{equation}
\mathbf{W}_{lp} = \boldsymbol{\alpha}_t^{\top} \mathbf{A}_{lp},
\end{equation}

Such a map demonstrates the global importance of every patch feature, regardless of where and which layer it is located at. This map is then used to modulate the original stacked features $\mathbf{F}_{\text{stack}}$ via element-wise multiplication, followed by a residual connection and layer normalization $LN(\cdot)$:

\begin{equation}
\mathbf{F}_{\text{lp}} = \text{LN}(\mathbf{F}_{\text{stack}} \odot \mathbf{W}_{lp} + \mathbf{F}_{\text{stack}}).
\end{equation}

where $\mathbf{F}_{\text{lp}} \in \mathbb{R}^{L \times (N \times d)}$ are the features aligned with the textual query from a joint patch-layer perspective. These features are then reshaped back to $\mathbb{R}^{L \times N \times d}$ to recover the per-layer structure for the next stage. With such an attention mechanism, we provide a more comprehensive textual-image alignment than only considering the patch- or layer-wise impact.

\subsection{Progressive Attention Integration}
\label{sec:unify}
Although LPWCA provides finer-grained attention, LWCA is crucial for focusing on semantically relevant layers, while PWCA constrains attention to consistent regions across layers. Without them, the learned features could be semantically or spatially inconsistent, making them difficult for humans to interpret. Therefore, Progressive Attention Integration (PAI) is proposed to integrate all three mechanisms harmoniously.

\textbf{Integration with LWCA.} 
Based on the globally-aligned features $\mathbf{F}_{\text{lp}}$ from LPWCA (reshaped to $L \times N \times d$), we apply a revised LWCA to provide continuous semantic attention. Specifically, the visual features are first spatially averaged to obtain a set of layer-level descriptors:

\begin{equation}
\mathbf{F}_{\text{layer}} = \left[\text{AvgPool}(\mathbf{F}^{1}_{\text{lp}}); \cdots; \text{AvgPool}(\mathbf{F}^{L}_{\text{lp}})\right] \in \mathbb{R}^{L \times d}.
\end{equation}

Then, cross-attention scores $\mathbf{A}_{l} \in \mathbb{R}^{T \times L}$ are computed between textual embeddings and layer descriptors, followed by aggregation using the same token importance weights $\boldsymbol{\alpha}_t$:

\begin{equation}
\mathbf{A}_{l} = \frac{1}{\sqrt{d}} \mathbf{Q}(\mathbf{F}_t) \mathbf{K}(\mathbf{F}_{\text{layer}})^{\top};
\end{equation}

\begin{equation}
\mathbf{w}_{l} = \boldsymbol{\alpha}_t^{\top} \mathbf{A}_{l}, \quad \mathbf{w}_{l} \in \mathbb{R}^{L}.
\end{equation}

Previous approaches to LWCA often select specific layers or cluster them to avoid sharp, noisy transitions in attention weights, which could disrupt the semantic smoothness across layers \cite{sung2023ecoflap,li2025instructionguidedfusionmultilayervisual,lin2025healthgpt}. However, this strategy risks discarding valuable information from the omitted layers. To address this, we introduce a Gaussian smoothing kernel applied to the raw layer attention scores $\mathbf{w}_{l}$. This method allows us to utilize information from all layers while simultaneously enforcing a smooth attention distribution, thus obtaining the final, refined layer weights $\hat{\mathbf{w}}_{l} \in \mathbb{R}^L$ that maintain both completeness of information and semantic consistency.

The semantically-aligned visual representation $\mathbf{F}_{\text{semantic}} \in \mathbb{R}^{N \times d}$ is derived via a weighted aggregation of the globally-aligned layer features. Let $\hat{\mathbf{F}}_{\text{lp}}$ be the result of the weighted sum. A residual connection and layer normalization are then applied:
\begin{equation}
\hat{\mathbf{F}}_{\text{lp}} = \sum_{l=1}^{L} \hat{w}_{l, l} \cdot \mathbf{F}^{l}_{\text{lp}}
\end{equation}
\begin{equation}
\mathbf{F}_{\text{semantic}} = \text{LN}\left(\hat{\mathbf{F}}_{\text{lp}} + \text{AvgPool}(\hat{\mathbf{F}}_{\text{lp}})\right).
\end{equation}

\textbf{Integration with PWCA.} Furthermore, to maintain regional consistency, we apply PWCA on $\mathbf{F}_{\text{semantic}}$. We first compute cross-attention between language tokens and the patch features of $\mathbf{F}_{\text{semantic}}$:

\begin{equation}
\mathbf{A}_{p} = \frac{1}{\sqrt{d}} \mathbf{Q}(\mathbf{F}_t) \mathbf{K}(\mathbf{F}_{\text{semantic}})^{\top},
\end{equation}

where $\mathbf{A}_{p} \in \mathbb{R}^{T \times N}$. The scores are then aggregated using the token importance weights $\boldsymbol{\alpha}_t$ to get patch weights $\mathbf{w}_{p} \in \mathbb{R}^{N}$:

\begin{equation}
\mathbf{w}_{p} = \boldsymbol{\alpha}_t^{\top} \mathbf{A}_{p}.
\end{equation}

Finally, a residual connection and layer normalization are applied to obtain the regionally-aligned visual representation $\mathbf{F}_{\text{regional}} \in \mathbb{R}^{N \times d}$:

\begin{equation}
\mathbf{F}_{\text{regional}} = \text{LN}\left( \mathbf{F}_{\text{semantic}} \odot (1 + \mathbf{w}_{\text{p}}) \right).
\end{equation}

To preserve both the original high-level visual semantics and the newly refined features, we concatenate $\mathbf{F}_{\text{regional}}$ with the original final-layer visual feature $\mathbf{F}^L_v$:
\begin{equation}
\mathbf{F}_{\text{fused}} = [\mathbf{F}_{\text{regional}}; \mathbf{F}^{L}_v] \in \mathbb{R}^{N \times 2d}.
\end{equation}

\textbf{Visual-textual Feature Fusion}
To align with the hidden dimension $d$ of the large language model, we apply a visual projection head $\text{Proj}_{\text{vis}}: \mathbb{R}^{2d} \rightarrow \mathbb{R}^{d}$ to each fused patch token. Subsequently, the resulting visual representation is then concatenated with the textual embeddings $\mathbf{F}_{t}$ and passed into a large language model for visual-language predictions (e.g., answer generation, captioning):

\begin{equation}
\hat{Y} = \text{LLM}([\mathbf{F}_{t};  \text{Proj}_{\text{vis}}(\mathbf{F}_{\text{fused}})]).
\end{equation}

Through such a progressive integration of LPWCA, LWCA, and PWCA, the final visual feature $\mathbf{F}_{\text{fused}}$ from PAI is tightly aligned with textual query, which supports the optimal performacne of VLM after the visual-textual feature fusion. Meanwhile,  it is also further constrained in semantic smoothness and regional consistency, which provides understandable attention map for human being.

The effectiveness of CCRA also depends on a few interpretable hyperparameters, such as the layer smoothing kernel size and embedding dimensions. Their impacts are discussed in Appendix \ref{app:sensitivity}.

\begin{algorithm}[t]
\caption{Training and Inference of CCRA-based Vision-Language Model}
\label{alg:ccra_vlm}
\KwIn{Image $I \in \mathbb{R}^{H \times W \times 3}$, Text sequence $\mathcal{T}$ with token length $T$, Task label $Y$ (for training)}
\KwOut{Prediction $\hat{Y}$ or updated model parameters}
\BlankLine
\textbf{1. Visual and Text Encoding} \\
$\mathbf{F}_{\text{stack}} \leftarrow \text{VisualEncoder}(I)$
$\mathbf{F}_{t} \leftarrow \text{TextEncoder}(\mathcal{T})$
\BlankLine
\textbf{2. Consistent Cross-layer Regional Alignment (CCRA)} \\
$\mathbf{F}_{\text{lp}} \leftarrow \text{LPWCA}(\mathbf{F}_{t}, \mathbf{F}_{\text{stack}})$
$\mathbf{F}_{\text{semantic}} \leftarrow \text{LWCA}(\mathbf{F}_{t}, \mathbf{F}_{\text{lp}})$
$\mathbf{F}_{\text{regional}} \leftarrow \text{PWCA}(\mathbf{F}_{t}, \mathbf{F}_{\text{semantic}})$
$\mathbf{F}_{\text{fused}} \leftarrow \text{Fuse}(\mathbf{F}_{\text{regional}}, \mathbf{F}^{L}_v)$
\BlankLine
\textbf{3. Training Stage 1: Feature Alignment Pretraining} \\
$\hat{Y} \leftarrow \text{LLM}([\mathbf{F}_{t};  \text{Proj}_{\text{vis}}(\mathbf{F}_{\text{fused}})])$ \tcp*{Caption prediction; freeze VisualEncoder and LLM}
$\mathcal{L}_{\text{pretrain}} \leftarrow \text{CrossEntropy}(\hat{Y}, Y)$
\BlankLine
\textbf{4. Training Stage 2: End-to-End Finetuning} \\
$\hat{Y} \leftarrow \text{LLM}([\mathbf{F}_{t}; \mathbf{F}_{\text{fused}}])$ \tcp*{Task-specific prediction; freeze VisualEncoder only}
$\mathcal{L}_{\text{finetune}} \leftarrow \text{CrossEntropy}(\hat{Y}, Y)$
\BlankLine
\textbf{5. Model Inference (if label $Y$ is not available)} \\
\textbf{Execute steps 1 and 2, then skip steps 3 and 4}
$\hat{Y} \leftarrow \text{LLM}([\mathbf{F}_{t}; \mathbf{F}_{\text{fused}}])$
\end{algorithm}

\begin{table*}[h]
\caption{Comparison across 10 benchmarks. Models are grouped by whether they adopt vision-language alignment.}
\setlength{\tabcolsep}{2pt}
\label{tab:full_comparison}
\resizebox{\textwidth}{!}{%
\begin{tabular}{@{}lccccccccccccc@{}}
\toprule
\multicolumn{1}{l|}{Model} & \multicolumn{1}{c|}{LLM} & \multicolumn{1}{c|}{Resolution} & \multicolumn{1}{c|}{Train Data} & \multicolumn{1}{c|}{GQA} & \multicolumn{1}{c|}{SQA} & \multicolumn{1}{c|}{TextVQA} & \multicolumn{1}{c|}{VizWiz} & \multicolumn{1}{c|}{MMB-en} & \multicolumn{1}{c|}{MM-Vet} & \multicolumn{1}{c|}{SEED-I} & \multicolumn{1}{c|}{MMMU} & \multicolumn{1}{c|}{MME-p} & POPE \\ \midrule
\multicolumn{1}{l|}{Metric} & \multicolumn{3}{l|}{} & \multicolumn{1}{l|}{Acc(\%)} & \multicolumn{1}{l|}{Acc(\%)} & \multicolumn{1}{l|}{Acc(\%)} & \multicolumn{1}{l|}{Acc(\%)} & \multicolumn{1}{l|}{Acc(\%)} & \multicolumn{1}{l|}{Acc(\%)} & \multicolumn{1}{l|}{Acc(\%)} & \multicolumn{1}{l|}{Acc(\%)} & \multicolumn{1}{l|}{MME-p Score} & \multicolumn{1}{l}{F1-Score} \\ \midrule
\multicolumn{14}{c}{\textbf{Models without Vision-Language Alignment}} \\ \midrule
\multicolumn{1}{l|}{LLaVA-v1.5-7B} & \multicolumn{1}{c|}{Vicuna-v1.5-7B} & \multicolumn{1}{c|}{336} & \multicolumn{1}{c|}{0.5M+0.6M} & \multicolumn{1}{c|}{61.9} & \multicolumn{1}{c|}{67.1} & \multicolumn{1}{c|}{58.1} & \multicolumn{1}{c|}{53.2} & \multicolumn{1}{c|}{63.9} & \multicolumn{1}{c|}{32.8} & \multicolumn{1}{c|}{67.2} & \multicolumn{1}{c|}{34.9} & \multicolumn{1}{c|}{1480.6} & 86.9 \\
\multicolumn{1}{l|}{LLaVA-v1.5-13B} & \multicolumn{1}{c|}{Vicuna-v1.5-13B} & \multicolumn{1}{c|}{336} & \multicolumn{1}{c|}{0.5M+0.6M} & \multicolumn{1}{c|}{63.3} & \multicolumn{1}{c|}{71.0} & \multicolumn{1}{c|}{61.3} & \multicolumn{1}{c|}{53.6} & \multicolumn{1}{c|}{67.7} & \multicolumn{1}{c|}{36.1} & \multicolumn{1}{c|}{68.2} & \multicolumn{1}{c|}{34.9} & \multicolumn{1}{c|}{-} & 87.2 \\
\multicolumn{1}{l|}{MiniGPT-v2} & \multicolumn{1}{c|}{LLaMA 2-7B} & \multicolumn{1}{c|}{448} & \multicolumn{1}{c|}{-} & \multicolumn{1}{c|}{60.1} & \multicolumn{1}{c|}{-} & \multicolumn{1}{c|}{-} & \multicolumn{1}{c|}{53.6} & \multicolumn{1}{c|}{9.4} & \multicolumn{1}{c|}{-} & \multicolumn{1}{c|}{-} & \multicolumn{1}{c|}{-} & \multicolumn{1}{c|}{-} & - \\
\multicolumn{1}{l|}{IDEFICS} & \multicolumn{1}{c|}{LLaMA-7B} & \multicolumn{1}{c|}{224} & \multicolumn{1}{c|}{1.6B} & \multicolumn{1}{c|}{38.4} & \multicolumn{1}{c|}{-} & \multicolumn{1}{c|}{25.9} & \multicolumn{1}{c|}{35.5} & \multicolumn{1}{c|}{48.2} & \multicolumn{1}{c|}{-} & \multicolumn{1}{c|}{-} & \multicolumn{1}{c|}{-} & \multicolumn{1}{c|}{-} & - \\
\multicolumn{1}{l|}{LLaMA-Adapter-v2} & \multicolumn{1}{c|}{LLaMA-7B} & \multicolumn{1}{c|}{336} & \multicolumn{1}{c|}{0.6M} & \multicolumn{1}{c|}{-} & \multicolumn{1}{c|}{-} & \multicolumn{1}{c|}{-} & \multicolumn{1}{c|}{-} & \multicolumn{1}{c|}{41.0} & \multicolumn{1}{c|}{31.5} & \multicolumn{1}{c|}{32.7} & \multicolumn{1}{c|}{29.8} & \multicolumn{1}{c|}{972.7} & - \\ \midrule
\multicolumn{14}{c}{\textbf{Models with Vision-Language Alignment}} \\ \midrule
\multicolumn{1}{l|}{DenseConnector} & \multicolumn{1}{c|}{-} & \multicolumn{1}{c|}{-} & \multicolumn{1}{c|}{-} & \multicolumn{1}{c|}{63.8} & \multicolumn{1}{c|}{69.5} & \multicolumn{1}{c|}{59.2} & \multicolumn{1}{c|}{-} & \multicolumn{1}{c|}{66.8} & \multicolumn{1}{c|}{32.7} & \multicolumn{1}{c|}{-} & \multicolumn{1}{c|}{34.8} & \multicolumn{1}{c|}{-} & 86.6 \\
\multicolumn{1}{l|}{MMFuser} & \multicolumn{1}{c|}{-} & \multicolumn{1}{c|}{-} & \multicolumn{1}{c|}{-} & \multicolumn{1}{c|}{62.8} & \multicolumn{1}{c|}{68.7} & \multicolumn{1}{c|}{58.8} & \multicolumn{1}{c|}{53.4} & \multicolumn{1}{c|}{67.5} & \multicolumn{1}{c|}{32.6} & \multicolumn{1}{c|}{60.8} & \multicolumn{1}{c|}{-} & \multicolumn{1}{c|}{1479.7} & 86.3 \\
\multicolumn{1}{l|}{IGVA} & \multicolumn{1}{c|}{Vicuna-v1.5-7B} & \multicolumn{1}{c|}{336} & \multicolumn{1}{c|}{0.5M+0.6M} & \multicolumn{1}{c|}{63.1} & \multicolumn{1}{c|}{70.2} & \multicolumn{1}{c|}{59.4} & \multicolumn{1}{c|}{54.3} & \multicolumn{1}{c|}{66.9} & \multicolumn{1}{c|}{33.5} & \multicolumn{1}{c|}{68.3} & \multicolumn{1}{c|}{36.4} & \multicolumn{1}{c|}{1519.8} & 87.8 \\
\multicolumn{1}{l|}{mPLUG-Owl2} & \multicolumn{1}{c|}{LLaMA 2-7B} & \multicolumn{1}{c|}{448} & \multicolumn{1}{c|}{384M+1.2M} & \multicolumn{1}{c|}{56.1} & \multicolumn{1}{c|}{68.7} & \multicolumn{1}{c|}{54.3} & \multicolumn{1}{c|}{54.5} & \multicolumn{1}{c|}{64.5} & \multicolumn{1}{c|}{36.2} & \multicolumn{1}{c|}{57.8} & \multicolumn{1}{c|}{-} & \multicolumn{1}{c|}{1450.2} & 86.2 \\
\multicolumn{1}{l|}{Qwen-VL-Chat} & \multicolumn{1}{c|}{Qwen-7B} & \multicolumn{1}{c|}{448} & \multicolumn{1}{c|}{1.4B+50M+0.3M} & \multicolumn{1}{c|}{57.5} & \multicolumn{1}{c|}{68.2} & \multicolumn{1}{c|}{61.5} & \multicolumn{1}{c|}{38.9} & \multicolumn{1}{c|}{60.6} & \multicolumn{1}{c|}{-} & \multicolumn{1}{c|}{65.4} & \multicolumn{1}{c|}{35.9} & \multicolumn{1}{c|}{1487.6} & - \\ \midrule
\multicolumn{14}{c}{\textbf{CCRA}} \\ \midrule
\multicolumn{1}{l|}{\textbf{Ours}} & \multicolumn{1}{c|}{Vicuna-v1.5-7B} & \multicolumn{1}{c|}{336} & \multicolumn{1}{c|}{0.5M+0.6M} & \multicolumn{1}{c|}{\textbf{64.2}} & \multicolumn{1}{c|}{\textbf{71.3}} & \multicolumn{1}{c|}{\textbf{63.1}} & \multicolumn{1}{c|}{\textbf{54.6}} & \multicolumn{1}{c|}{\textbf{67.9}} & \multicolumn{1}{c|}{\textbf{37.5}} & \multicolumn{1}{c|}{\textbf{69.6}} & \multicolumn{1}{c|}{\textbf{37.6}} & \multicolumn{1}{c|}{\textbf{1525.6}} & \textbf{88.9} \\ \bottomrule
\end{tabular}%
}
\end{table*}

\section{Experiment}
We begin by introducing the experimental setup in Section~\ref{sec:exp_setup}. To comprehensively assess the task-oriented effectiveness of CCRA, we evaluate the model on ten widely used benchmarks, which test various capabilities of vision-language models, such as compositional reasoning, OCR, instruction following, and domain-specific understanding. We compare CCRA with a suite of state-of-the-art models to demonstrate its superior performance. In addition, to evaluate interpretability of CCRA, we visualize feature attention heatmaps to qualitatively illustrate the improved vision-language consistency achieved by our model (Section~\ref{sec:results}). Finally, we conduct a detailed ablation study to validate the necessity and effectiveness of each component within the CCRA framework (Section~\ref{sec:Ablation}).

\subsection{Experimental Setup}
\label{sec:exp_setup}
Our training procedure follows the two-stage strategy of LLaVA-v1.5-7B. In the pre-training stage, we train on the LLaVA-LCS-558K dataset, which includes 558K image–text pairs, where we do not apply CCRA herein as annotations are not available. In the instruction-tuning stage, we switch to the LLaVA-Instruct-665K dataset, which contains 665K multimodal instruction samples, and CCRA is applied to optimize the VLM's vision-language consistency to improve performance and interpretability. To highlight the improvement introduced by CCRA compared to LLaVA, the overall optimization strategy is almost entirely inherited from LLaVA. More details can be found in Appendix \ref{app:exp set}.
\begin{figure*}[h]
\centering
\includegraphics[width=0.7\textwidth]{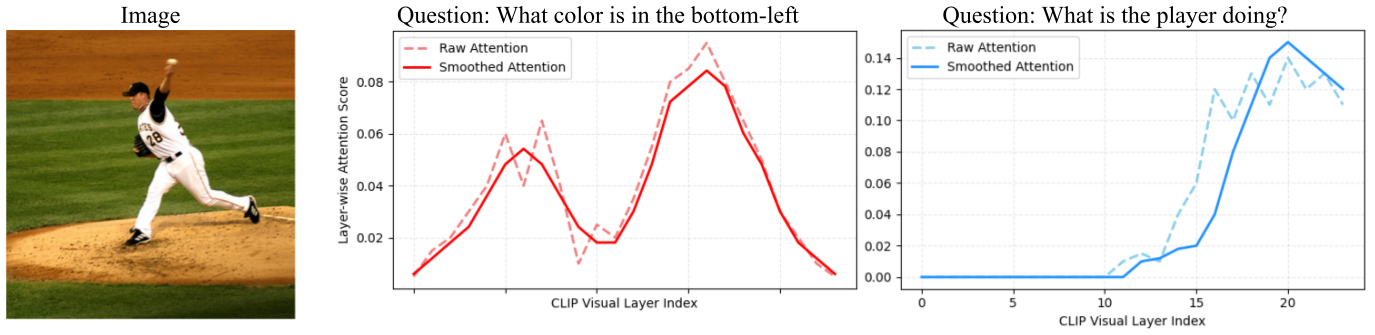}
\caption{
Comparison of LWCA distributions for queries of different semantic levels. 
Shallow appearance-based queries activate earlier layers, while high-level reasoning queries attend to deeper layers. 
Smoothed attention curves (solid) reveal more coherent trends than raw attention (dashed).
}
\label{fig:layer_attention_distribution}
\end{figure*}
\begin{figure}[h]
\centering
\includegraphics[width=\columnwidth]{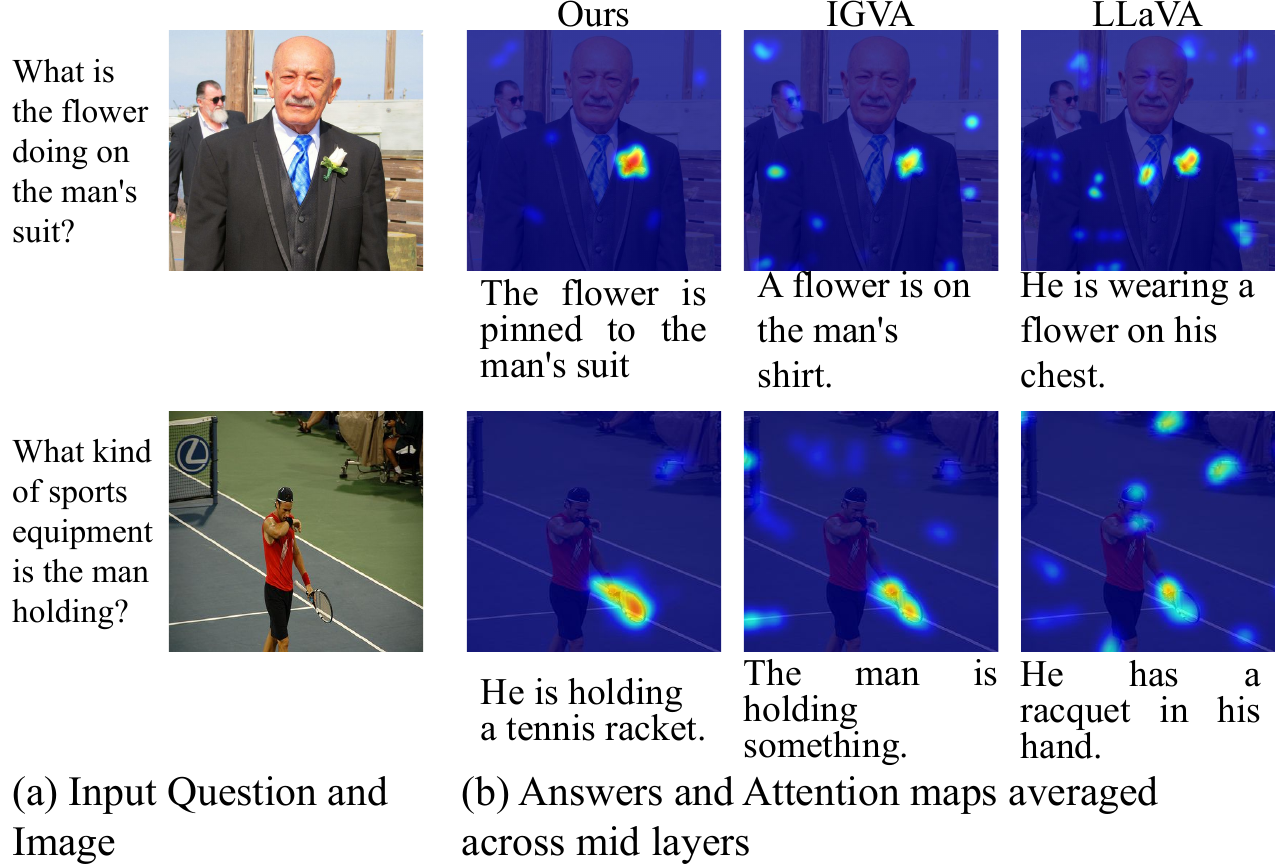}
\caption{
Cross-attention visualization of language tokens attending to visual patches, averaged over mid-layers of the LLM. 
CCRA (leftmost) exhibits more consistent and focused attention compared to grouped-layer (IGVA) fusion or fixed-layer (LLaVA) baselines. 
Below each attention map, we additionally present the model’s textual answer, providing a qualitative link between attention precision and language output.
}
\label{fig:llm_attention_vis}
\end{figure}
We evaluate CCRA on ten widely-used benchmarks with diverse task requirements, they are GQA~\cite{hudson2019gqa}, ScienceQA(SQA)~\cite{lu2022learn}, TextVQA~\cite{singh2019towards}, VizWiz~\cite{gurari2018vizwiz}, MMB-en~\cite{liu2024mmbench}, MM-Vet~\cite{yu2023mm}, SEED-I~\cite{li2024seed}, MMMU~\cite{yue2024mmmu}, MME-p~\cite{fu2023mme}, and POPE~\cite{li2023evaluating}, Notably, for ScienceQA, we only evaluate on the set with image context. More details on dataset can be found in Appendix \ref{dataset}. Parallelly, SOTA methods including LLaVA-v1.5-7B~\cite{liu2024improved}, LLaVA-v1.5-13B~\cite{liu2024improved}, mPLUG-Owl2~\cite{ye2024mplug}, MiniGPT-v2~\cite{chen2023minigpt}, LLaMA-Adapter-v2~\cite{gao2023llama}, IDEFICS~\cite{laurenccon2023obelics}, DenseConnector~\cite{yao2024dense}, MMFuser~\cite{cao2024mmfuser}, IGVA~\cite{li2025instruction} and Qwen-VL-Chat~\cite{bai2023qwen}, are used to compare with CCRA, to demonstrate CCRA's advance.

\subsection{Results and Analysis}
\label{sec:results}

We reported the performance of CCRA and competitors on these ten tasks in Table~\ref{tab:full_comparison}. The metric used is accuracy except for MME-p, which uses the MME-P score, and POPE, which uses the F-1 score, as per their respective guidelines~\cite{fu2023mme, yao2023pope}. The competitors are grouped according to whether they incorporate a vision-language alignment mechanism.

In general, CCRA outperforms all competitors on all ten benchmarks, which highlights its effectiveness in optimizing task-oriented performance. Compared to LLaVA-v1.5-7B (our baseline), CCRA provides contrastively better performance on TextVQA, where fine-grained OCR and understanding capabilities are needed. This benefit is provided by the harmonious integration of attentions in CCRA. Notably, MME-p, where object location and number are assessed, and POPE, where model robustness against hallucination is evaluated, also require very robust and accurate vision-language alignment. CCRA also demonstrates significant improvement over LLaVA in these two benchmarks, highlighting its effectiveness in adaptively aligning to diverse task queries. Notably, the attention operations only increased 3.55M parameters, which is neglectable compared to the 7B size of the LLaVA, but it empowered LLaVA-7B to realize superior performance than LLaVA-V1.5-13B.

Figure \ref{fig:layer_attention_distribution} intuitively illustrates how semantic layer attention adaptively aligns with the task query in CCRA. We tested two distinctive task queries: “What color is in the bottom-left corner?” (a low-level appearance query) and “What is the baseball player doing?” (a high-level reasoning query). As it shows, the appearance query activates the middle layers (e.g., layers 5–8), which capture low-level appearance features such as color and texture, while the reasoning query yields a sharp attention shift toward deeper layers (e.g., layers 16–22), which tackle semantics and pose understanding. Notably, thanks to our Gaussian smoothness design, local oscillations and inconsistent transitions between adjacent layers in the raw attention are filtered, while information from all layers is maintained.

Table~\ref{tab:full_comparison} also quantitatively demonstrates that vision-language alignment is vital in improving task-orientedperformance, as models that adopted it generally perform better than those that do not. It should be highlighted that even when compared to models that consider vision-language alignment, our model still achieves the best results on most benchmarks. Specifically, it outperforms IGVA by +1.1 on GQA, +1.3 on SEED-I, and +1.0 on MMB-en; and surpasses MMFuser by +4.3 on TextVQA, and +4.9 on MM-Vet. These improvements can be attributed to our key designs: LPWCA and PAI, which further optimize the alignment by introducing layer-patch correlation and a harmonious integration of diverse attentions.

To intuitively demonstrate the contribution of CCRA to reasoning improvement, we visualize the cross-attention maps from layers 12 to 20 of the LLM after text-patch embedding fusion (Figure\ref{fig:llm_attention_vis}), as these layers focus on attention from each text token to visual embeddings. The visualization shows that CCRA empowers the LLM to focus more precisely on task-relevant image regions (e.g., object attributes, fine-grained cues), resulting in more accurate and semantically aligned responses. In contrast, recent IGVA and LLaVA models cannot provide highly consistent attention, which corresponds to their vague or partially incorrect answers.

\begin{figure*}[h]
\centering
\includegraphics[width=0.9\linewidth]{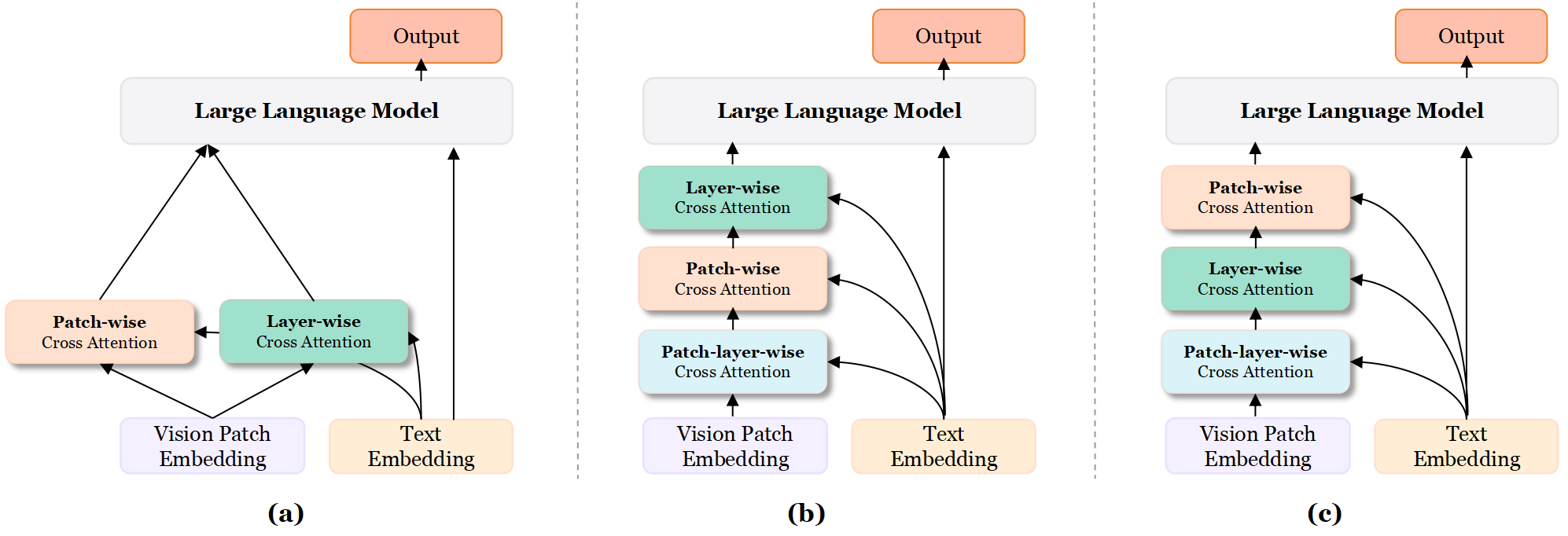}
\caption{
Comparison of cross-attention coordination strategies for combining patch-level (regional) and layer-level (semantic) information. 
\textbf{(a)} \textbf{Decoupled Integration}: Patch-wise (PWCA) and layer-wise (LWCA) attentions are applied independently and fused in parallel before LLM input.
\textbf{(b)} \textbf{Shuffled Integration}: A reversed ordering of patch-wise and layer-wise attention compared to PAI.
\textbf{(c)} \textbf{Proposed PAI (Patch-layer-wise Attention Integration)}: Our progressive strategy that sequentially applies patch-layer-wise, layer-wise, and patch-wise attention to iteratively refine visual features.
}
\label{fig:patch_layer_model_variants}
\end{figure*}

Additionally, Figure \ref{fig:layer_attention_distribution} and Figure \ref{fig:llm_attention_vis} also provide clear interpretable information at semantic and regional levels, respectively. This comes from the dedicated design of PAI, as it progressively adds Gaussian-smoothed LWCA and PWCA to further constrain semantic smoothness and region consistency, which alleviates the mismatch between attentions. More details can be found in Appendix \ref{app:exp result}.

\begin{table}[h]
\centering
\caption{Ablation study results of CCRA}
\setlength{\tabcolsep}{2pt}
\label{tab:vqa_ablation}
\resizebox{1\columnwidth}{!}{%
\begin{tabular}{@{}c|c|c|c|c|c|c|c|c|c|c@{}}
\toprule
Model Variant & GQA & SQA & TextVQA & VizWiz & MMB-en & MM-Vet & SEED-I & MMMU & MME-p & POPE \\ \midrule
w/o Patch-wise & 62.9 & 69.5 & 58.8 & 53.2 & 67.1 & 36.8 & 68.3 & 36.4 & 1513.2 & 87.5 \\
w/o Layer-wise & 63.4 & 70.1 & 58.2 & 53.9 & 66.7 & 36.0 & 68.1 & 36.9 & 1510.3 & 87.0 \\
w/o Layer-patch-wise & 62.7 & 70.6 & 59.0 & 52.8 & 66.9 & 37.1 & 68.7 & 36.3 & 1514.6 & 88.7 \\
w/o Gaussian smoothing & 63.8 & 70.9 & 59.1 & 54.0 & 67.3 & 36.5 & 69.0 & 36.7 & 1516.2 & 88.2 \\ \midrule
\textbf{Ours} & \textbf{64.2} & \textbf{71.3} & \textbf{63.1} & \textbf{54.6} & \textbf{67.9} & \textbf{37.5} & \textbf{69.6} & \textbf{37.6} & \textbf{1525.6} & \textbf{88.9} \\ \bottomrule
\end{tabular}
}
\end{table}

\subsection{Ablation study}
\label{sec:Ablation}
To assess the contribution of each attention mechanism in our framework, we conduct ablation studies on these benchmarks, as shown in Table~\ref{tab:vqa_ablation}. The results clearly demonstrate that all three attention modules—patch-wise, layer-wise, and layer-patch-wise—play essential and complementary roles in enhancing performance. Removing PWCA leads to noticeable performance drops on tasks such as TextVQA, VizWiz, and MMMU, which depend heavily on fine-grained regional understanding. Excluding LWCA results in lower scores on benchmarks like SQA, MMB-en, and SEED-I, where semantic abstraction is crucial. Omitting LPWCA degrades performance on MM-Vet and POPE, indicating its importance in maintaining regional-semantic consistency across layers.

We further evaluate the effect of the Gaussian smoothing applied to the raw layer attention scores in LWCA (shown in Table~\ref{tab:vqa_ablation}). After removing this step, a moderate performance drop occurs across most benchmarks, especially those requiring stable cross-layer reasoning (e.g., SQA, POPE). This confirms that smoothing helps suppress sharp and noisy transitions in the attention distribution, thus maintaining semantic continuity as Figure~\ref{fig:layer_attention_distribution} shown. 

\begin{table}[h]
\centering
\setlength{\tabcolsep}{2pt}
\caption{Performance comparison between PAI and its variants}
\label{tab:coordination_strategy_comparison}
\resizebox{1\columnwidth}{!}{%
\begin{tabular}{@{}c|c|c|c|c|c|c|c|c|c|c@{}}
\toprule
Model Variant & GQA & SQA & TextVQA & VizWiz & MMB-en & MM-Vet & SEED-I & MMMU & MME-p & POPE \\ \midrule
Decoupled Integration & 63.1 & 70.2 & 58.7 & 53.0 & 67.0 & 36.5 & 68.5 & 36.6 & 1497.7 & 87.4 \\
Shuffled Integration & 63.8 & 70.8 & 59.3 & 53.7 & 67.3 & 37.0 & 68.9 & 36.8 & 1501.6 & 88.2 \\
\textbf{Ours (PAI)} & \textbf{64.2} & \textbf{71.3} & \textbf{63.1} & \textbf{54.6} & \textbf{67.9} & \textbf{37.5} & \textbf{69.6} & \textbf{37.6} & \textbf{1525.6} & \textbf{88.9} \\ \bottomrule
\end{tabular}
}
\end{table}

To better understand why PAI is superior to other integration strategies, we compared PAI (Figure~\ref{fig:patch_layer_model_variants} (c)) with two of its variants (Figure~\ref{fig:patch_layer_model_variants}): (a) \textbf{Decoupled Integration}, where patch- and layer-wise cross attentions are fused in parallel before entering the LLM; and (b) \textbf{Shuffled Integration}, which reverses the order of patch- and layer-wise operations in PAI.

\begin{figure}[h]
\centering
\includegraphics[width=1\linewidth]{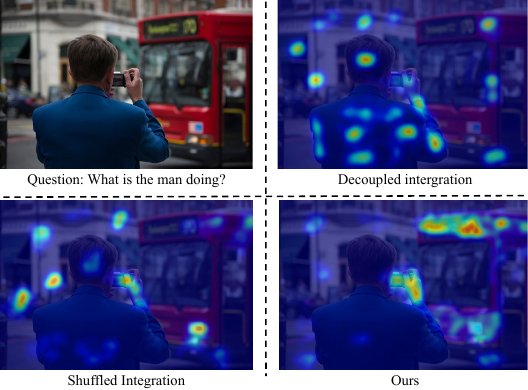}
\caption{
Cross-attention maps from the LLM under different coordination strategies. 
CCRA (c) yields sharper and more semantically aligned attention compared to (a) independent coordination and (b) a variant progressive design. 
These visualizations confirm the effectiveness of our full progressive attention refinement.
}
\label{fig:attention_map_model_variants}
\end{figure}

The performance is reported in Table~\ref{tab:coordination_strategy_comparison}, where our original PAI demonstrates the best performance. Furthermore, the text-attention from layers 12-20 is visualized in Figure~\ref{fig:attention_map_model_variants} to intuitively check the vision-language alignment. The figure shows that the proposed PAI produces sharper, more consistent, and semantically focused attention, while both variants exhibit dispersed or misaligned patterns. These qualitative results align well with the quantitative gains, further proving the effectiveness of our dedicated design for PAI.

\FloatBarrier
\section{Conclusion}
We proposed Consistent Cross-layer Regional Alignment (CCRA), a novel framework that harmonically integrates diverse attention mechanisms in Vision Language Models through two key innovations. Layer-Patch-wise Cross Attention (LPWCA) captures fine-grained regional-semantic correlations by jointly considering spatial and depth-wise semantics, providing more comprehensive text-image alignment than existing patch-wise or layer-wise mechanisms alone. Progressive Attention Integration (PAI) systematically coordinates all three attention mechanisms in sequence, ensuring consistency from semantic to regional levels while maximizing individual attention benefits and preventing attention drift. Experimental results across ten benchmarks demonstrate that our CCRA-enhanced model achieves state-of-the-art performance, outperforming all baseline methods while adding only 3.55M parameters, with visualization analyses confirming more focused and semantically aligned attention patterns.

Beyond the advanced performance of CCRA, we realize that there is still an optimization space to be explored. Future work could explore alternative smoothing mechanisms for optimizing LWCA beyond the current Gaussian smoothing approach to achieve better integration within the progressive attention framework and further improve vision-language alignment performance.

\newpage
\bibliographystyle{abbrv}
\bibliography{references}

\newpage
\appendix
\appendix
\section{Dataset Overview}
\label{dataset}
Detailed information about all our 10 datasets is introduced as follows.

\textbf{GQA}
GQA \cite{hudson2019gqa} is a large-scale benchmark for compositional visual reasoning, comprising real-world images annotated with structured scene graphs. It contains over 22 million questions across 100K images, designed to evaluate multi-step reasoning that involves attributes, spatial relations, and logical consistency.

\textbf{SQA}
The Spatial Question Answering (SQA) \cite{hu2019you} dataset evaluates a model's ability to comprehend spatial relationships between objects. It features questions derived from the VQA dataset, with a specific focus on prepositions such as “left of” or “above.”

\textbf{TextVQA}
TextVQA \cite{singh2019textvqa} evaluates the ability to answer questions that require recognizing and interpreting text within images. The dataset includes over 45K questions on images with naturally occurring text, requiring robust OCR and multimodal reasoning capabilities.

\textbf{VizWiz}
VizWiz \cite{gurari2018vizwiz} contains visual questions posed by visually impaired users. The dataset consists of real-world, often noisy images and naturally phrased questions, establishing it as a benchmark for accessibility-focused VQA and open-ended question understanding.

\textbf{MMBench-en}
MMBench-en \cite{li2023mmbench} is a comprehensive benchmark that evaluates English-language multimodal models across multiple tasks. It features classification-style questions covering perception, reasoning, and instruction-following to assess model generalization across diverse domains.

\textbf{MM-Vet}
MM-Vet \cite{fu2023mmvet} is a curated benchmark that probes vision-language alignment and reasoning consistency. It includes challenging visual questions with distractors, evaluating a model's ability to perceive images accurately rather than relying on language priors. MM-Vet employs a normalized score, computed by averaging accuracies across seven core visual reasoning skills, each weighted equally.

\textbf{SEED-I}
SEED-I \cite{zhu2023seed}, a subset of SEED-Bench, evaluates multimodal instruction-following in vision-centric tasks. It includes open-ended prompts that require image understanding, captioning, and grounding based on natural user instructions.

\textbf{MMMU}
The Massive Multi-discipline Multimodal Understanding (MMMU) benchmark \cite{zeng2023mmmu} contains over 10K expert-level questions spanning 57 disciplines, such as medicine, law, and physics. It assesses the capacity of models to handle college-level, multi-disciplinary questions.

\textbf{MME-p}
MME-p \cite{fu2023mme}, the perception-oriented subset of the MME benchmark, evaluates fine-grained visual capabilities, including object counting, attribute recognition, OCR, and positional reasoning. It is designed to isolate core perceptual skills from higher-level cognitive reasoning.

\textbf{POPE}
The POPE benchmark \cite{yao2023pope} evaluates whether models ground their answers in visual input or rely on spurious language biases. It presents contrastive image-question pairs to assess the strength of visual grounding and robustness against hallucination.

\section{Detailed Experimental Setup}
\label{app:exp set}

In this appendix, we provide a detailed overview of our two-stage training pipeline, covering both optimization settings and resource utilization.

\paragraph{Training Configurations.}
The complete hyperparameters and hardware configurations for both training stages are summarized in Table~\ref{tab:config_vertical}. In the pre-training stage, we train the model on the LLaVA-LCS-558K dataset for one epoch (2,179 steps), using a batch size of 256 and a learning rate of $1 \times 10^{-3}$, consistent with the LLaVA-v1.5-7B training strategy. For the visual instruction tuning stage, we switch to the LLaVA-Instruct-665K dataset, reduce the batch size to 128, and decrease the learning rate to $1 \times 10^{-5}$ to ensure training stability. Both stages employ a cosine learning rate schedule with a 3\% warm-up period and the AdamW optimizer. We conduct distributed training on 8 NVIDIA A100 GPUs (80GB each), with DeepSpeed ZeRO-2 optimization enabled for memory efficiency.

\paragraph{Resource Usage.}
As detailed in Table~\ref{tab:config_vertical}, the pre-training stage is memory-efficient, as only a small adapter module is trained while the remainder of the model remains frozen. In contrast, the instruction tuning stage involves optimizing the full model, including the language model, vision encoder, and our proposed CCRA module. This full-model optimization leads to an increase in peak GPU memory usage, despite a reduction in batch size. The CCRA module contributes minimally to this increase, adding less than 3GB of memory overhead. Even with 32-bit precision, the entire model can be accommodated on a single A100 80GB GPU. With the exception of the CCRA integration, the training pipeline strictly adheres to the original LLaVA setup to ensure a fair and controlled comparison.

\begin{table}[]
\centering
\scriptsize
\setlength{\tabcolsep}{2pt}
\caption{Training configuration and resource usage per stage.}
\label{tab:config_vertical}
\begin{tabular}{@{}c|c|c@{}}
\toprule
\textbf{Setting} & \textbf{Pretraining} & \textbf{Visual Instruction Tuning} \\ \midrule
Dataset & LLaVA‑LCS‑558K & LLaVA‑Instruct‑665K \\
GPUs & 8×A100 80GB & 8×A100 80GB \\
Batch size & 256 & 128 \\
Total steps & 2,179 & 5,194 \\
Epochs & 1 & 1 \\
Learning rate & 1e‑3 & 1e‑5 \\
Schedule & cosine & cosine \\
Warm‑up & 3\% & 3\% \\
Optimizer & AdamW & AdamW \\
ZeRO Stage & ZeRO‑2 & ZeRO‑2 \\
Wall‑clock time (h) & 4 & 10 \\
Aggregate GPU‑hours & 32 & 80 \\
Peak VRAM / card & 73GB & 76GB \\ \bottomrule
\end{tabular}
\end{table}

\section{Additional Experimental Results}
\label{app:exp result}

\subsection{Qualitative Visualization of Cross Attention}
To better understand how our proposed Consistent Cross-layer Regional Alignment (CCRA) framework utilizes multi-layer visual features under language guidance, we visualize the attention behaviors of its three core components: Layer-wise Cross Attention, Layer-patch-wise Cross Attention, and Patch-wise Cross Attention. For visualization, we select representative layers for LWCA and LPWCA to illustrate their contributions across depth and spatial dimensions. In contrast, LWCA operates on the fused representation generated by the previous two modules, and its visualization reflects spatial alignment after semantic refinement.

\begin{figure}[h]
\centering
\includegraphics[width=1\linewidth]{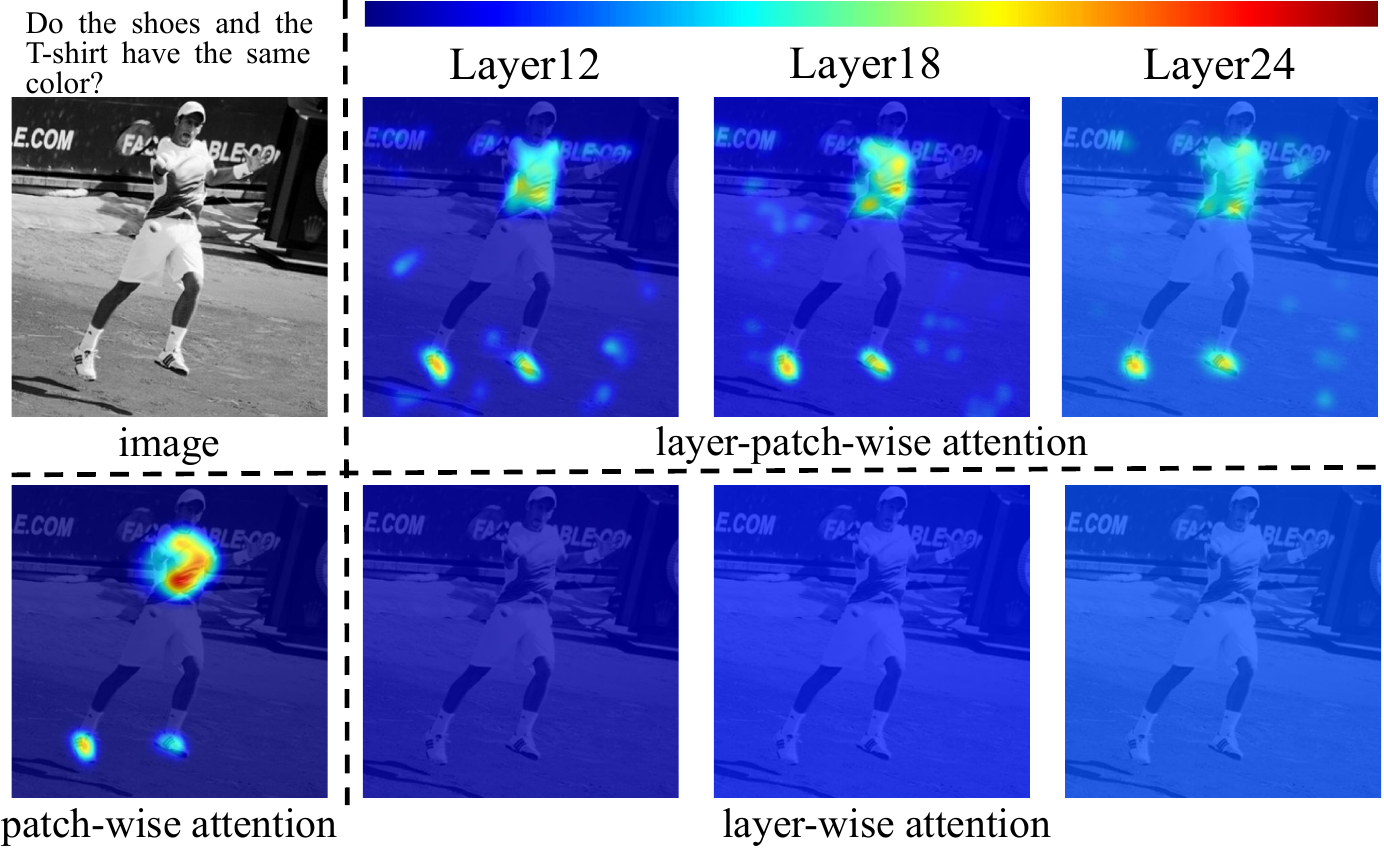}
\caption{
Visualization of attention maps from the three modules in our CCRA framework, given the question “Do the shoes and the T-shirt have the same color?”. 
}
\label{fig:module_attention_map}
\end{figure}

As illustrated in Figure~\ref{fig:module_attention_map}, we present visualizations from the three attention modules in our CCRA framework, conditioned on the question “Do the shoes and the T-shirt have the same color?” The figure is structured in a 4×2 layout. The top-left image shows the input and the corresponding question. To its right, the three attention maps correspond to Layer-patch-wise Cross Attention applied to Layer 12, 18, and 24, respectively. This module jointly considers spatial and depth-wise semantics, enabling the model to simultaneously select the most relevant regions and visual layers. The strong focus on the player’s shirt and shoes exemplifies the module’s ability to align visual content with the question. The bottom-left cell visualizes the output of Patch-wise Cross Attention, which operates on the fused multi-layer representation. This module captures fine-grained spatial alignment after layer aggregation, refining the visual grounding with enhanced localization precision. The three maps on the bottom row (right) correspond to Layer-wise Cross Attention at Layer 12, 18, and 24. This mechanism estimates the relative token-wise importance within each layer, offering semantic filtering along the feature hierarchy but without explicit spatial fusion. Together, these visualizations showcase how each attention module contributes distinct and complementary capabilities—joint spatial-depth selection, fine-grained spatial alignment, and layer-level semantic emphasis—which collectively enhance the interpretability and cross-modal alignment of our model.

\subsection{Hyperparameter Sensitivity Analysis}
\label{app:sensitivity}

To evaluate the robustness of CCRA to key hyperparameters, we conduct a sensitivity analysis on the attention projection dimension, $d_{\text{hidden}}$, and the Gaussian smoothing kernel size, $k$. These parameters respectively control the expressiveness of text-image alignment and the continuity of layer-wise semantic attention.

\begin{figure}[h]
    \centering
    \includegraphics[width=0.9\linewidth]{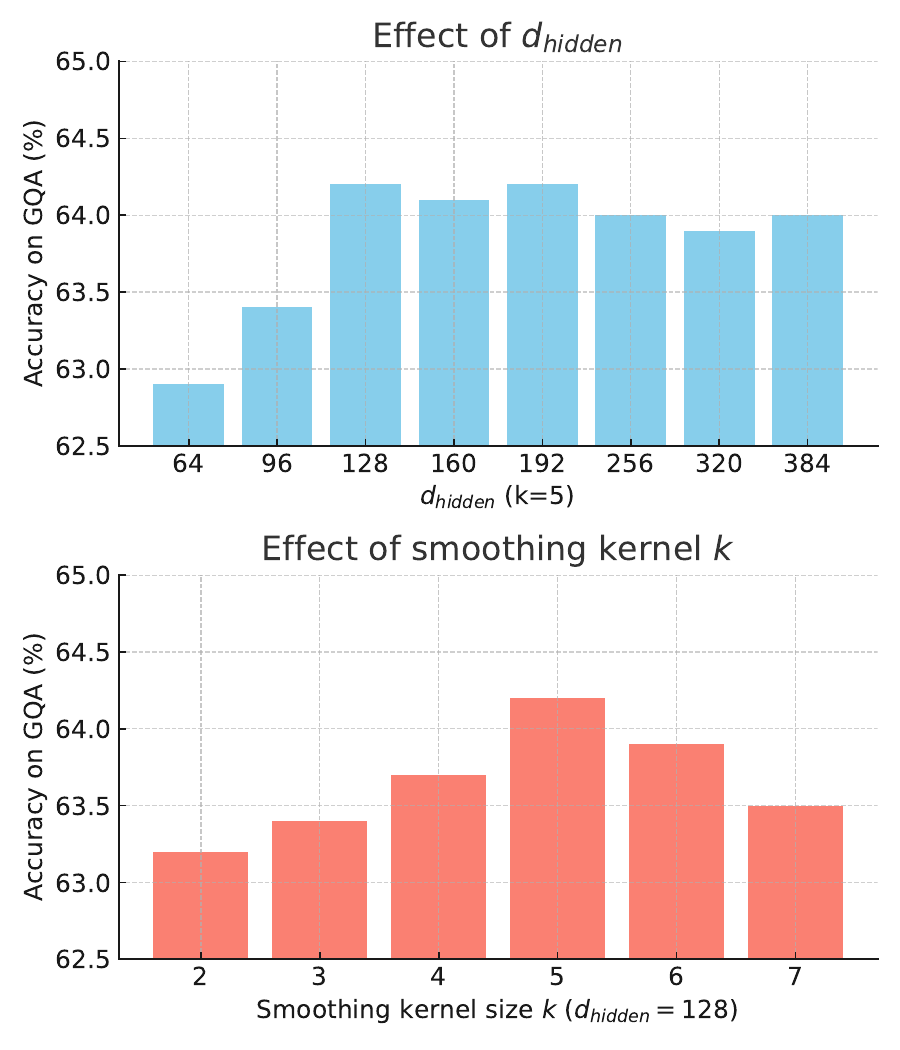}
    \caption{Sensitivity of CCRA on the GQA dataset with respect to hidden dimension $d_{\text{hidden}}$ and smoothing kernel size $k$.}
    \label{fig:hyper_sensitivity}
\end{figure}

We vary $d_{\text{hidden}}$ across the set $\{64, 96, 128, 160, 192, 256, 320, 384\}$ and $k$ across $\{2, 3, 4, 5, 6, 7\}$, while evaluating accuracy on the GQA benchmark. As shown in Figure~\ref{fig:hyper_sensitivity}, the model demonstrates consistent performance across these settings. Notably, $d_{\text{hidden}}=128$ and $k=5$ yield the optimal trade-off between performance and stability, aligning with our default configuration.

\section{Qualitative Examples of Model Predictions}
\label{app:qualitative_examples}
To better illustrate the behavior of our CCRA model across diverse visual-language tasks, we present qualitative examples categorized into four groups: counting, OCR, object grounding, and binary (yes/no) questions. Each example includes an input image, a task-specific question, and the model's generated response.
\begin{figure*}[h]
    \centering
    \includegraphics[width=0.7\linewidth]{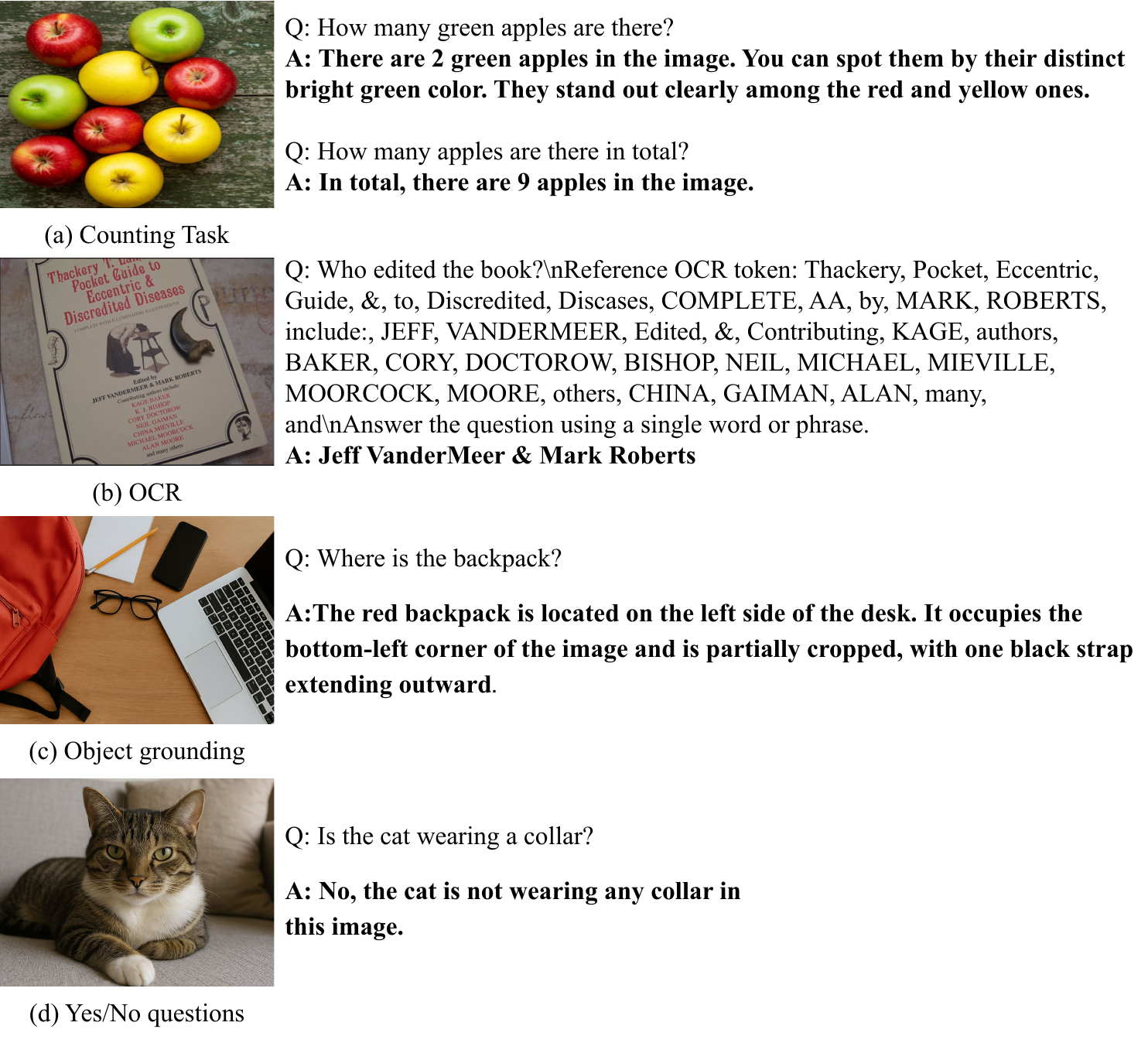}
    \caption{Qualitative examples of CCRA model predictions across diverse tasks: (a) Counting (e.g., number of apples), (b) OCR (e.g., book editors), (c) Object Grounding (e.g., backpack location), and (d) Yes/No questions (e.g., presence of a collar).}
    \label{fig:qualitative_examples}
\end{figure*}
\FloatBarrier
 As shown in Figure~\ref{fig:qualitative_examples}, the model demonstrates fine-grained understanding across various challenges, such as distinguishing apple colors and quantities, reading book titles and editor names, spatially grounding objects in cluttered scenes, and answering binary attribute-based questions. These results highlight the model's capability to effectively align vision and language under varying semantic demands.


\end{document}